\documentclass{llncs}

\usepackage{soul}
\usepackage{url}
\usepackage[hidelinks]{hyperref}
\usepackage[utf8]{inputenc}
\usepackage[small]{caption}
\usepackage{graphicx}
\usepackage{amsmath}
\usepackage{amssymb}
\usepackage{booktabs}
\usepackage{algorithm}
\usepackage{algorithmic}
\newcommand{\hidden}[1]{}

\usepackage{eurosym}

\newcommand{\cftual}{\Box\!\!\!\rightarrow}

\newcommand{\game}{\mathcal{G}}

\newcommand{\fh}{\hat{f}}

\newcommand{\play}{\rho}

\newcommand{\x}{{\hat{x}}}

\newtheorem{rem}{Remark}
\newtheorem{cor}{Corollary}

\newenvironment{sketch}{\vspace{1ex}\noindent{\it Proof sketch.}\hspace{0.5em}}
	{\hfill\qed\vspace{1ex}}

\usepackage{linguex}

\title{Fair and Adequate Explanations}

\author{Nicholas Asher\inst{1} \and
Soumya Paul\inst{2} \and
Chris Russell\inst{3}}
\authorrunning{N. Asher et al.}
\institute{IRIT, Universit\'e Paul Sabatier, 118 route de Narbonne, 31062 Toulouse, France
\email{asher@irit.fr}\\
\and
Telindus, 18 rue du Puits Romain, L-8070 Bertrange, Luxembourg\\
\email{soumya.paul@telindus.lu}
\and
Amazon Research, T\"ubingen, Germany\\
\email{ cmruss@amazon.com}}

\begin{document}

\maketitle

\vspace{-0.2cm}

\date{}

\vspace{-0.2cm}

\begin{abstract}
Recent efforts have uncovered various methods for providing explanations that can help interpret the behavior of machine learning programs. 
Exact explanations with a rigorous logical foundation provide valid and complete explanations, but they have an epistemological problem: they may be too complex for humans to understand and too expensive to compute even with automated reasoning methods.  
Interpretability requires {\em good} explanations that humans can grasp and can compute.  

We take an important step toward specifying what good explanations are by analyzing the epistemically accessible and pragmatic aspects of explanations.  We characterize sufficiently good, or fair and adequate, explanations in terms of counterfactuals and what we call the {\em conundra of the explainee}, the agent that requested the explanation.  We provide a correspondence between logical and mathematical formulations for counterfactuals to examine the partiality of counterfactual explanations that can hide biases; we define fair and adequate explanations in such a setting.  We then provide formal results about the algorithmic complexity of fair and adequate explanations.

\end{abstract}
\section{Introduction} 
Explaining the predictions of sophisticated machine-learning algorithms is an important issue for the foundations of AI. 
Recent efforts \cite{ribeiro:etal:2016,ribeiro:etal:2018,wachter:etal:2017,marques-silva:etal:2019,bachoc:etal:2018} have shown various methods for providing explanations.  Among these, model-based, logical approaches that completely characterise one aspect of the decision promise complete and valid explanations. 

Such logical methods are thus {\em a priori} desirable, but they have an epistemological problem: they may be too complex for humans to understand or even to write down in human-readable form. 
Interpretability requires {\em epistemically accessible} explanations, explanations humans can grasp {\em and compute}.  Yet what is a sufficiently complete and {\em adequate} epistemically accessible explanation, a {\em good explanation} still needs analysis \cite{murdoch:etal:2019}.   We propose to characterize sufficiently good, or fair and adequate, explanations in terms of counterfactuals---explanations, that is that are framed in terms of what would have happened had certain conditions (that do not obtain) been the case---and what we call the {\em conundrum} and {\em fairness requirements} of the {\em explainee}, the person who requested the explanation or for whom the explanation is intended).  It is this conundrum that makes the explainee request an explanation.  Counterfactual explanations, as we argue below, are a good place to start for finding accessible explanations, because they are typically more compact than other forms of explanation.

 We argue that a fair and adequate explanation is relative to the cognitive constraints and fairness requirements of an  explainee $\cal{E}$ \cite{bromberger:1962,achinstein:1980,miller:2019}.  $\cal{E}$ asks for an explanation for why $\pi$ when she wasn't expecting $\pi$.  Her not expecting $\pi$ follows from beliefs that must now be revised---how to specify this revision is the conundrum of $\cal{E}$.  An adequate explanation is a pragmatic act that should solve the conundrum that gave rise to the request for explanation; solving the conundrum makes the explanation useful to ${\cal E}$ \cite{holzinger:etal:2020}.  In addition, an adequate explanation must lay bare biases that might be unfair or injurious to $\cal{E}$ (the fairness constraint).  In effect, this pragmatic act is naturally modelled in a game theoretic setting in which the explainer must understand explainee E’s conundrum and respond so as to resolve it.  A cooperative explainer will provide an explanation in terms of the type he assigns to E, as the type will encode the relevant portions of E’s cognitive state. On the other hand the explainee will need to interpret the putative explanation in light of her model of the explainer’s view of his type. Thus, both explainer and explainee have strategies that exploit information about the other---naturally suggesting a game theoretic framework for analysis. 
 
 In developing our view of fair and adequate explanations, we will exploit both the logical theory of counterfactuals \cite{lewis:1973} and mathematical approaches for adversarial perturbation techniques \cite{kusner:etal:2017,younes:2018,peyre:2019,kurakin:etal:2016,dube:2018,bachoc:etal:2018}.   We provide a correspondence between logical and mathematical formulations for counterfactuals, and we analyze how counterfactual explanations can hide biases.  We then formalize conundra and fair and adequate explanations, and we develop a game theoretic setting for proving computational complexity results for finding fair and adequate explanations in non cooperative settings.


\section{Background on explanations}
Following \cite{bromberger:1962,achinstein:1980}, we take explanations to be answers to {\em why} questions.  Consider the case where a bank, perhaps using a machine learning program, judges $\cal{E}$'s application for a bank loan and $\cal{E}$ is turned down.   ${\cal E}$ is in a position to ask a {\em why} question like, 
\ex. why was I turned down for a loan? \label{question}

when her beliefs would  not have predicted this.  Her beliefs might not have been sufficient to infer that she wouldn't get a loan; or her beliefs might have been mistaken---they might have led her to conclude that she would get the loan.  In any case, ${\cal E}$ must now revise her beliefs to accord with reality.   {\em Counterfactual explanations}, explanations expressed with counterfactual statements, help ${\cal E}$ do this by offering an incomplete list of relevant factors that together with unstated properties of ${\cal E}$ entail the {\em explanandum}---the thing ${\cal E}$ needs explained, in this case her not getting the loan.   For instance, the bank might return the following answer to \ref{question}:

\ex. \label{bankfact} Your income is  \euro 50K per year.  

\ex. \label{bankcounterfactual} If your income had been \euro 100K  per year, you would have gotten the loan.

\noindent
The counterfactual statement \ref{bankcounterfactual} states what given all of ${\cal E}$'s other qualities would have been sufficient to get the loan.  But since her income is in fact not \euro 100K  per year, the semantics of counterfactuals entails that ${\cal E}$ does not get the loan.  \ref{bankcounterfactual} also proposes to ${\cal E}$ how to revise her beliefs to make them accord with reality, in that it suggests that she mistakenly thought that her actual salary was sufficient for getting the loan and that the correct salary level is \euro 100K  per year.\footnote{\cite{fan:toni:2015a} provide a superficially similar picture to the pragmatic one we present, but their aim is rather different, to provide a semantics for argumentation frameworks.  For us the pragmatic aspect of explanations is better explained via a game theoretic framework; see below.}  

Counterfactual explanations, we have seen, are {\em partial}, because they do not explicitly specify logically sufficient conditions for the prediction.  They are also {\em local}, because their reliance on properties of a particular sample makes them valid typically only for that sample.   Had we considered a different individual, say ${\cal D}$, the bank's explanation for their treatment of ${\cal D}$ might have differed.  ${\cal D}$ might have had different, relevant properties from ${\cal E}$; for instance, ${\cal D}$ might be just starting out on a promising career with a salary of \euro  50K per year, while ${\cal E}$ is a retiree with a fixed income.   

The partiality and locality of counterfactuals make them simpler and more epistemically accessible than other forms of explanation.  Moreover, the logical theory of counterfactuals enables us to move from a counterfactual to a  complete and logically valid explanation. So in principle counterfactual explanations can provide both rigour and epistemic accessibility.  But not just any partiality will do, since partiality makes possible explanations that are misleading, that hide injurious or unfair biases.  
To show how the partiality of counterfactual explanations can hide unfair biases, consider the following scenario.    
The counterfactual in (2) might be true but it also might be {\em misleading}, hiding an unfair bias.   (1)-(2) can be true while another, more morally repugnant explanation that hinges on $\cal{E}$'s being female is also true. Had $\cal{E}$ been male, she would have gotten the loan with her actual salary of \euro 50K per year.  A fair and adequate explanation should expose such biases.

We now move to a more abstract setting.  Let $\hat{f} \colon X^n \rightarrow Y$ be a machine learning algorithm, with  $X^n$ an n-dimenstional feature space encoding data and $Y$ the prediction space.  Concretely, we assume that $\hat{f}$ is some sort of classifier.  When $\hat{f} = \pi$, an explainee may want an explanation, an answer to the question,``why $\pi$?"  We will say that an explanation is an event by an {\em explainer}, the provider of the explanation,  directed towards the explainee (the person requesting the explanation or to whom the explanation is directed) with a conundrum.  An explanation will consist of of an {\em explanandum}, the event or prediction to be explained, an {\em explanans}, the information that is linked in some way to the explanandum so as to resolve the  explainee's conundrum.  When the explanation is about a particular individual, we call that individual the {\em focal point} of the explanation.

Explanations have thus several parameters.  The first is the scope of the explanation.  For a  {\em global} explanation of $\hat{f}$, the explainee  wants to know the behavior of $\hat{f}$ over the total space $X^n$. But such an explanation may be practically uncomputable; and for many purposes, we might only want to know how $\hat{f}$ behaves on a selection  of data points of interest or focal points, like $\cal{E}$'s bank profile in our example.\footnote{We are implicitly assuming that $\hat{f}$ is too complex or opaque for its behaviour to be analyzed statically.}  Explanations that are restricted to focal points are {\em local} explanations.  

Explanations of program behavior also differ as to the nature of the {\em explanans}.  
In this paper, we will be concerned with {\em external} explanations that involve an explanatory link between features of input or feature space $X$ and the output in $Y$ without considering any internal states of the learning mechanism \cite{friedrich:zanker:2011}.  These are attractive epistemically, because unpacking the algorithms' internal states and assigning them a meaning can be a very complicated affair.  

A third pertinent aspect of explanations concerns the link between {\em explanans} and the {\em explanandum}. \cite{hempel:1965,marques-silva:etal:2019,ignatiev:etal:2020} postulate a deductive or logical consequence link between {\em explanans} and {\em explanandum}.  \cite{marques-silva:etal:2019} represent $\hat{f}$ as a set of logic formulas ${\cal M}(\hat{f})$.  By assuming features with binary values\footnote{By increasing the number of literals we can simulate non binary values, so this is not really a limitation as long as the features are finite.}, an {\em instance} is then a set of literals that assigns values to every feature in the feature space.  An {\em abductive} explanation of why $\pi$ is a subset minimal set of literals ${\cal I}$ such that  ${\cal M}(\hat{f}), {\cal I} \models \pi$. 
Abductive explanations exploits universal generalizations and a deductive consequence relation.  They explain why {\em any instance} $\x$ that has ${\cal I}$ is such that $\fh(\x) = \pi$ and hence are known as {\em global} explanations \cite{molnar:2019}.

Counterfactuals offer a natural way to provide epistemically accessible, partial explanations of properties of individuals or focal points. The counterfactual in \ref{bankcounterfactual} gives a sufficient reason for $\cal{E}$'s getting the loan, {\em all other factors of her situation being equal} or being as equal as possible ({\em ceteris paribus}) given the assumption of a different salary for $\cal{E}$.   Such explanations are often called {\em local} explanations \cite{doshi:etal:2017,molnar:2019}, as they depend on the nature of the focal point; they are also partial \cite{wachter:etal:2017}, because the antecedent of a counterfactual are not by themselves logically sufficient to yield the formula in the consequent.  Deductive explanations, on the other hand, are invariant with respect to the choice of focal point.  But because counterfactual explanations exploit {\em ceteris paribus} conditions, factors that deductive explanations must mention can remain implicit in a counterfactual explanation.   Thus, counterfactual explanations are typically more compact and thus in principle easier to understand.\footnote{See \cite{ignatiev:etal:2020} for some experimental evidence of this.}  Counterfactuals are also intuitive vehicles for explanations as they also encode an analysis of causation \cite{lewis:1973}.


\noindent
\noindent

\subsection{Counterfactual explanations for learning algorithms}

The canonical semantics for a counterfactual language ${\cal L}$, which is a propositional language to which a two place modal operator $\cftual$ is added, as outlined in \cite{lewis:1973} exploits a possible worlds model for propositional logic, 
$\mathfrak{A} = \langle W, \leq, [\![ .]\!] \rangle$, where: $W$ is a non-empty set (of worlds), $\leq$ is a ternary similarity relation ($w' \leq_{w} w''$), and $[\![ .]\!]: P \rightarrow W \rightarrow \{0, 1\}$ assigns to elements in $P$, the set of proposition letters or atomic formulas of the logic, a function from worlds to truth values or set of possible worlds.  Then, where 
$\models$ represents truth in such a model, we define truth recursively as usual for formulas of ordinary propositional logic and for counterfactuals $\psi \ \cftual \phi$, we have:
\begin{definition} \label{semantics}
  $\mathfrak{A}, w \ \models \psi \ \cftual \phi$  just in case:  $\forall w',
   \mbox{if } \mathfrak{A}, w' \models \psi \mbox{ and } \forall w'' (\mathfrak{A}, w'' \models \psi \rightarrow  w'\leq_w w''), \mbox{ then: } \mathfrak{A}, w' \models \phi$ 
\end{definition}
What motivates this semantics  with a similarity relation?   We can find both epistemic and metaphysical motivations.  Epistemically, finding a closest or most similar world in which the antecedent $\phi$ of the counterfactual $\phi \ \cftual \psi$ is true to evaluate its consequent $\psi$ follows a principle of belief revision \cite{gardenfors:makinson:1988}, according to which it is rational to make minimal revisions to one's epistemic state upon acquiring new conflicting information.  A metaphysical motivation comes from the link Lewis saw between counterfactuals and causation; $\neg \phi \cftual \neg \psi$ implies that if $\phi$ hadn't been the case, $\psi$ wouldn't have been the case, capturing much of the semantics of the statement $\phi$ {\em caused} $\psi$.  The truth of such intuitive causal statements, however, relies on the presence of a host of secondary or enabling conditions.  Intuitively the statement that if I had dropped this glass on the floor, it would have broken is true; but in order for the consequent to hold after dropping the glass, there are many elements that have to be the same in that counterfactual situation as in the actual world---the floor needs to be hard, there needs to be a gravitational field around the strength of the Earth's that accelerates the glass towards the floor, and many other conditions.  In other words, in order for such ordinary statements to be true, the situation in which one evaluates the consequent of a counterfactual has to resemble very closely the actual world.

Though intuitive, as this logical definition of counterfactuals stands, it is not immediately obvious how to apply it to explanations of learning algorithm behavior.  We need to adapt it to a more analytical setting.    
We will do so by interpreting the similarity relation appealed to in the semantics of counterfactuals as a distance function or norm as in \cite{williamson:1988} over the feature space $X^n$, an n-dimensional space, used to describe data points.  To fill out our semantics for counterfactuals in this application, we identify instances in $X^n$ as the relevant ``worlds'' for the semantics of the counterfactuals. We now need to specify a norm for $X^n$.  A very simple norm assumes that each dimension of $X^n$ is orthogonal and has a Boolean set of values; in this case, $X^n$ has a natural $L_1$ norm or Manhattan or edit distance \cite{salzberg:1991}.\footnote{In fact, we only assume a finite set of finitely valued features, since an n-valued feature is definable with n Boolean valued features.   By complicating the language and logic \cite{deraedt:etal:2020}, we can have probability estimates on literals and so encode continuous feature spaces.}  While this assumption commits us to the fact that the dimensions of $X^n$ capture {\em all the causally relevant} factors and that they are all independent---both of which are false for typical instances of learning algorithms, it is simple and makes our problem concrete.  We will indicate below when our results depend on this simplifying assumption.

A logic of counterfactuals can now exploit the link between logic formulas, features of points in $X^n$, and a learning algorithm $\hat{f}$ described in \cite{marques-silva:etal:2019,karimi:etal:2020}.   Suppose a focal point $\hat{x}$ is such that $\hat{f}(\hat{x}) = \eta$.   A counterfactual $A \ \cftual \pi$ that is true at the point $\hat{x}$, where $\pi$ is a prediction incompatible with $\eta$, has an antecedent that is a conjunction of literals, each literal defining a feature value, and that provides a sufficient and minimal shift in the features of $\hat{x}$ to get the prediction $\pi$.   Each counterfactual that explains the behavior of $\hat{f}$ around a focal point $\hat{x} \in X^n$ thus defines a minimal transformation of the features of $\hat{x}$ to change the prediction.  We now define the transformations on  $X^n$ that counterfactuals induce.

\hidden{
\begin{definition}\label{def:trans} A transformation $\Delta$ is a function $\Delta: X^n \rightarrow X^n$. Given $x\in X^n$, and $\hat{f}(x) = \eta$, we shall be interested in the following types of transformations.
  \begin{enumerate}
  \item[(i)] $\Delta(x)$ is appropriate if $\hat{f}(\Delta(x)) = \pi$ where $\eta$ and $\pi$ are two incompatible predictions in $Y$.
  \item[(ii)] $\Delta(x)$ is minimally appropriate if it is appropriate and in addition, $\forall x' \in X$ such that $\hat{f}(x') = \pi$, $\|x' - x\|_{X^n} \geq \|\Delta(x) - x\|_{X^n}$, where $\|.\|_{X^n}$ is a natural norm on $X^n$.
  \end{enumerate}
\end{definition}

We also look at transformations that are restricted to a fixed set of indices.}

\begin{definition}\label{def:transfixed} Let $i\subset n$. A {\sf fixed transformation} $\Delta_i$ is a function $\Delta_i: X^n \rightarrow X^n$ such that for $x\in X^n$, if $\Delta_i(x) = y$, then $x$ and $y$ differ only in the dimensions in $i$.  We write $x =_i x'$ to mean that $x$ and $x'$ share the same values along dimensions $i$. Given $x\in X^n$, and $\hat{f}(x) = \eta$ and where  $\|.\|_{X^n}$ is a natural norm on $X^n$, we shall be interested in the following types of transformations.
  \begin{enumerate}
  \item[(i)] $\Delta_i(x)$ is {\sf appropriate} if $\hat{f}(\Delta_i(x)) = \pi$ where $\eta$ and $\pi$ are two incompatible predictions in $Y$.
  \item[(ii)] $\Delta_i(x)$ is {\sf minimally appropriate} if it is appropriate and in addition, $\forall x' \in X$ such that $\Delta_i(x)  =_i x'$ and $\hat{f}(x') = \pi$, $\|x' - x\|_{X^n} \geq \|\Delta_i(x) - x\|_{X^n}$.
  \item[(iii)] $\Delta_i(x)$ is {\sf sufficiently appropriate} if it is appropriate and in addition, for any $j\subsetneq i$, $\Delta_j(x)$ is not appropriate.
    \item[(iv)] $\Delta_i(x)$ is {\sf sufficiently minimally appropriate} if it is both sufficiently and minimally appropriate.
  \end{enumerate}
\end{definition}

\noindent 
Note that when $X$ is a space of Boolean features, then conditions (ii) and (iv) of Def. \ref{def:transfixed} trivially hold. Given a focal point $\hat{x}$ in $X^n$, minimally appropriate transformations represent the minimal changes necessary to the features of $\hat{x}$ to bring about a change in the value predicted by $\hat{f}$. 

Let $\fh: X^n \rightarrow Y$ and consider now a counterfactual language ${\cal L}_{\fh}$ with a set of formulas $\Pi$ that  describe the predictions in $Y$ of $\fh$.  
\begin{definition} \label{model:cftual}
A {\sf counterfactual model} ${\cal C}_{X^n, \hat{f}}$ for ${\cal L}_{\fh}$ with $\hat{f} \colon X^n \rightarrow Y$ is a triple $\langle W, \leq, [\![ .]\!] \rangle$ with $W$ a set of worlds $W = X^n$,  $\leq$ defined by a norm $||.||$ on $X^n$ and $[\![ .]\!]: P \cup \{\Pi\} \rightarrow W \rightarrow \{0, 1\}$ such that for $A \in P$, $[\![A]\!]_w = 1$ iff $w$ has feature $A$ and for $\pi \in \Pi$, 
$[\![ \pi]\!]_w = 1$ iff $\hat{f}(w) = \pi$.
\end{definition} 
Given  a counterfactual model ${\cal C}_{X^n, \hat{f}}$ for ${\cal L}_{\fh}$ with  norm $||.||$ on  $X^n$, we say that $||.||$ is ${\cal L}_{\fh}$ {\sf definable}  just in case for worlds $w, w_1 \in X^n$, there is a formula $\phi$ of ${\cal L}_{\fh}$ that {\sf separates} $w_1$ from  all $w_2 \in X^n$  such that  $\|w_2- w\| < \|w_1- w\|$---i.e. for all $w_2$,  $\|w_2- w\| < \|w_1- w\|$,   ${\cal C}_{X^n, \hat{f}}, w_1 
\models \phi$ and ${\cal C}_{X^n, \hat{f}}, w_2 
\not\models \phi$.  
\begin{proposition}.  \label{trans-cftual} Let  $\hat{f} \colon X^n \rightarrow Y$ and let ${\cal C}_{X^n, \hat{f}}$ be a counterfactual model for ${\cal L}_{\fh}$ with an ${\cal L}_{\fh}$ {\sf definable} norm.  Suppose also that $\hat{f}(w) = \eta$.  Then: \\
${\cal C}_{X^n, \hat{f}}, w \ \models \ \phi \ \cftual \pi$, where $\pi \in \Pi$ and $\phi$ is a separating formula iff there is a minimally appropriate transformation, $\Delta_i: X^n \rightarrow X^n$, where $\hat{f}(\Delta_i(w)) = \pi$, and ${\cal C}_{X^n, \hat{f}}, \Delta_i(w) \ \models \ A$.
\end{proposition}
\noindent
Proposition \ref{trans-cftual} follows easily from Definitions \ref{semantics}, \ref{def:transfixed} and  \ref{model:cftual}.   

Proposition \ref{trans-cftual} is general and can apply to many different norms and languages.  We will mostly be concerned here with a special and simple case: 
\begin{cor} \label{L1trans} Let ${\cal L}_{\fh}$ be a propositional language with a set $P$ of propositional letters, where $P$ is the set of Boolean valued features of $X^n$, and let ${\cal C}_{X^n, \hat{f}}$ be a counterfactual model for ${\cal L}_{\fh}$ with an $L_1$ norm. Then: \\${\cal C}_{X^n, \hat{f}}, w \ \models \ A \ \cftual \pi$, where $\pi \in \Pi$ and $A$ is a conjunction of literals in $P$ iff there is a minimally appropriate transformation over the dimensions $i$ fixed by $A$, $\Delta_i: X^n \rightarrow X^n$, where $\hat{f}(\Delta_i(w)) = \pi$, and ${\cal C}_{X^n, \hat{f}}, \Delta_i(w) \ \models \ A$.
\end{cor}

We can generate minimally appropriate transformations via efficient (poly-time) techniques like optimal transport or diffeomorphic deformations \cite{younes:2018,peyre:2019,kurakin:etal:2016,dube:2018,bachoc:etal:2018} for computing adversarial perturbations \cite{kusner:etal:2017}.  In effect all of these diverse methods yield counterfactuals or sets of counterfactuals given Proposition \ref{trans-cftual}.  
A typical definition of an adversarial perturbation of an instance $x$, given a classifier, is that it is a smallest change  to $x$ such  that  the  classification changes.  Essentially,  this is  a counterfactual by a different name. Finding a closest possible world to $x$ such that the classification changes is, under the right choice of distance function, the same as finding the smallest change to $x$ to get the classifier to make a different prediction.\footnote{Such minimal perturbations may not reflect the ground truth, the causal facts that our machine learning algorithm is supposed to capture with its predictions, as noted by \cite{laugel:etal:2019}.  We deal with this in Section 4.}  

The great advantage of Proposition \ref{trans-cftual} is that marries efficient techniques to generate counterfactual explanations with the logical semantics of counterfactuals that provides logically valid (LV) explanations from counterfactual explanations, unlike heuristic methods \cite{ribeiro:etal:2018,lundberg:lee:2017,ribeiro:etal:2018}.   Thus, counterfactual explanations build a bridge between logical rigour and computational feasibility.
\begin{proposition} \label{cftual-proof} A counterfactual explanation given by a minimally appropriate $\Delta_i(\x)$ in ${\cal C}_{X^n, \hat{f}}$, with an $L_1$ norm and $X^n$ with Boolean valued features, yields a minimal, LV explanation in at worst a linear number of calls to an NP oracle.
\end{proposition}
\noindent
\begin{sketch}The atomic diagram \cite{chang:keisler:1973}  of ${\cal C}_{X^n, \hat{f}}$ in which each world is encoded as a conjunction of literals (Boolean values of the features $P$ of $X^n$ together with predictions from $Y$), encodes ${\cal M}(\hat{f})$.  Further, given  Corollary \ref{L1trans} and Definition \ref{model:cftual}, each minimally appropriate $\Delta_i$ defines a set of literals ${\cal L}_{\Delta_i}$ describing $\Delta_i(\x)$ such that $\Delta_i(\x), {\cal M}(\hat{f}) \models \pi$. \cite{marques-silva:etal:2019,ignatiev:etal:2020} provide an algorithm for finding a subset minimal set of literals ${\cal E} \subseteq {\cal L}_{\Delta_i}$ with  ${\cal E}, {\cal M}(\hat{f}) \models \pi$ in a linear number relative to $|{\cal L}_{\Delta_i}|$ of calls to an NP oracle \cite{junker:2004}.
\end{sketch}


\section{From partial to more complete explanations}

We have observed that counterfactual explanations are intuitively simpler than deductive ones, as  
they typically offer only a partial explanation.
 In fact there are three sorts of partiality in a counterfactual explanation.  First, a counterfactual explanation is {\em deductively incomplete}; it doesn't specify the {\em ceteris paribus} conditions and so doesn't specify what is necessary for a proof of the prediction $\pi$  for a particular focal point.  Second, counterfactual explanations are also partial in the sense that they don't specify all the sufficient conditions that lead to $\pi$; they are hence {\em globally incomplete}. Finally, counterfactuals are partial in a third sense; they are also {\em locally incomplete}.  To explain this sense, we need a notion of {\em overdetermination}.
 \begin{definition} \label{def:over} A prediction $\pi \in Y$ by $\fh : X \rightarrow Y$ is {\sf overdetermined} for a focal point $\hat{x} \in X$ if the set of minimally sufficiently appropriate transformations of $\hat{x}$ 
 $$O(\x, \pi, \fh) = \{\Delta_i : \Delta_i(\hat{x}) \text{ is minimally sufficiently appropriate}\}$$
   contains at least two elements. 
\end{definition}
Locally incomplete explanations via counterfactuals can occur whenever $\hat{f}$'s counterfactual decisions are over-determined for a given focal point.   Many real world applications like our bank loan example will have this feature.  

Locally incomplete explanations can, given a particular ML model ${\cal M}_{\fh}$, hide implicitly defined properties that show $\fh$ to be unacceptably biased in some way and so pose a problem for fair and adequate explanations.  Local incompleteness allows for several explanatory counterfactuals with very different {\em explanans} to be simultaneously true.  This means that even with an explanation, $\hat{f}$ may act in ways unknown to the agent $\cal{E}$ or the public that is biased or unfair.  Worse, the constructor or owner of $\hat{f}$ will be able to conceal this fact if the decision for $\cal{E}$ is overdetermined, by offering counterfactual explanations using maps $\Delta$ that don't mention the biased feature.  
\begin{definition}
A {\sf prejudicial factor} $P$ is a map,  $P\colon X^n \rightarrow X^n$ and  $\hat{f}$ exhibits a {\sf biased dependency} on prejudicial factor $P$  just in case for some $i \neq 0$, $\Delta_i$, and for some incompatible predictions $\eta$ and $\pi$,
\begin{equation*}
\hat{f}(\x) = \hat{f}(\Delta_i(\x)) = \eta \ \mbox{ and } \
\hat{f}(P(\x)) = \hat{f}(P(\Delta_i(\x))) = \pi
\end{equation*}
 \end{definition}

\noindent
Dimensions of the feature space that are atomic formulas in ${\cal L}_{\hat{f}}$ can provide examples of a prejudicial factor $P$.  But prejudical factors $P$ may be also implicitly definable  in $M_{\hat{f}}$.  Assume that $\hat{.}$ is a map from real individuals $x$ to their representation as data points $\hat{x} \in \hat{X}$. Then: $P$ is $M_{\hat{f}}$ {\sf implicitly definable} just in case: for all $x$ such that $\x \in  \hat{X}$, $x  \in \|P\|$ iff for some boolean combination $E$ of atoms of ${\cal L}_{\hat{f}}$, $M_{\hat{f}} \models E(\hat{x})$.

We've just described some pitfalls of locally incomplete counterfactual explanations. We now show how to move from a partial picture of the behavior of $\hat{f}$ to a more complete one using counterfactuals.   Imagine that at a focal point $\hat{x}$, $\hat{f}(\hat{x}) = \eta$ and we want to know why not $\pi$.  
\begin{definition} In a counterfactual model ${\cal C}_{X^n, \hat{f}}$ with a set of Boolean valued features $P$, the collection of counterfactuals ${\bf S}_{{\cal C},\x, \pi} = \{ \phi \ \cftual \pi : {\cal C}_{X^{n}, \hat{f}}, \x \models \phi \ \cftual \pi \mbox{\it{ with }} \phi \mbox{\it{ a Boolean combination of values for atoms in }} P \}$  true at $\hat{x}$ gives the {\sf complete explanation} for why $\pi$ would have occurred at $\x$.
\end{definition}  

Appropriate transformations $\Delta_i$ on $X^n$ in a counterfactual model ${\cal C}_{X^n, \hat{f}}$ to produce $\pi$ associated with counterfactuals via Proposition \ref{trans-cftual} can capture ${\bf S}_{{\cal C}, \x, \pi} $ and permit us to plot the local complete explanation of $\hat{f}$ around a focal point $\hat{x}$ with regard to  prediction $\pi$.   
\begin{definition}
${\bf B}_{{\cal C}, \x, \pi}  = \{\Delta_i(\x) : \Delta_i$  is a minimal appropriate transformation for some $ i \subset n\}$ 
\end{definition}
\begin{proposition}
In a counterfactual model ${\cal C}_{X^n, \hat{f}}$ ,  ${\bf B}_{{\cal C}, \x, \pi} = \{ y \in X^n : \ \exists \ (\phi \ \cftual \psi) \ \in  {\bf S}_{{\cal C}, \x, \pi}  \mbox{\it{ such that }} y  \mbox{\it{ is a closest $\phi$ world to }} \x  \mbox{\it{ where }} {\cal C}_{X^n, \hat{f}}, y \models \psi \}. $
\end{proposition}
For the remainder of this section we will fix a counterfactual model ${\cal C}_{X^n, \hat{f}}$ to simplify notation.

We are interested in the space ${\cal N}_{\hat{f},\hat{x}, \pi}$ around $\x$ with boundary ${\bf B}_{\x, \pi}$.
\begin{definition}
\begin{enumerate}
\item ${\cal N}_{\fh, \x, \pi}$ is the subspace of $X^n$ such that (i) $\x \in {\cal N}_{\fh, \x, \pi}$ and (ii) ${\cal N}_{\fh,\pi, \x}$ includes in its interior all those points $z$ for which $\hat{f}(z) = \hat{f}(\x)$ and (iii) 
the boundary of ${\cal N}_{\fh, \x, \pi}$ is given by ${\bf B}_{\x, \pi}$.
\item  ${\cal N}^d_{\fh, \pi, \x} $ is a subspace of ${\cal N}_{\fh, \x, \pi} $ with boundary ${\bf B}^d_{\x, \pi}$, where ${\bf B}^d_{\x, \pi} = {\bf B}_{\x, \pi}  \cap B_d(\x)$, where $B_d(\x) = \{ y \in X^n: \|y - \x\| \leq d\}.$
\item  ${\bf S}^d_{\x, \pi} =  \{y 
\colon \exists (\phi \ \cftual \psi) \ \in  {\bf S}_{\x, \pi}   \wedge  {\cal C}_{X^n, \hat{f}}, y \models \psi \wedge \|y - \x \| \leq d \}.$
\end{enumerate}
\end{definition}

The set ${\bf S}_{\x, \pi} $ can have a complex structure in virtue of the presence of {\em ceteris paribus} assumptions.  Because strengthening of the antecedent fails in semantics for counterfactuals, 
the counterfactuals in \ref{nested} relevant to our example of Section 2 are all satisfiable at a world without forcing the antecedents of  \ref{nested}b or \ref{nested}c to be inconsistent:
\ex. \label{nested}
\a. If I were making \euro 100K euro, I would have gotten the loan.
\b. If I were making \euro 100K or more but were convicted of a serious financial fraud, I would not get the loan.
\c. If I were making \euro 100K or more and were convicted of a serious financial fraud but then the conviction was overturned and I was awarded a medal, I would get the loan.

\noindent
The closest worlds in which I make \euro 100k do not include a world $w$ in which I make \euro 100k but am also convicted of fraud.  Counterfactuals share this property with other conditionals that have been studied in nonmonotonic reasoning \cite{ginsberg:1986,pearl:1990}.  However, if the actual world turns out to be like $w$, then by weak centering \ref{nested}a turns out to be false, because the {\em ceteris paribus} assumption in \ref{nested}a is that the actual world is one in which I'm not convicted of fraud. 

In ${\bf S}_{\x, \pi} $  we can count how many times the value of the consequent changes as we move from one antecedent to a logically more specific one (e.g., does the prediction flip from $A$ to $A \wedge C$ or from $A \wedge C$ to $A \wedge C \wedge D$).   For generality, we will also include in the number of flips, the flips that happen when we change the Boolean value of a feature---going from $A$ to $\neg A$ for example.   We will call the number of flips the {\em flip degree} of ${\bf S}_{\x, \pi} $.  

There is an important connection between the flip degree of ${\bf S}_{\x, \pi}$ and the geometry of ${\cal N}_{\fh, \x, \pi}$.  In a counterfactual model, the move from one antecedent $\phi_1$ of a counterfactual $c_1$ a to logically more specific antecedent $\phi_2$ of $c_2$, with $c_1, c_2 \in {\bf S}_{\x, \pi} $ will, given certain assumptions about the underlying norm yield $\x < y < z$, with $y$ being a closest to $\x$ point verifying $\phi_1$ and $z$ a closest point verifying $\phi_2$.  In fact we generalize this property of norms.
\begin{definition}
A norm $||.||$ in a counterfactual model   ${\cal C}_{X^n, \hat{f}}$ {\sf respects the logical specificity} of the model  iff for any $z \in X^n$ such that ${\cal C}_{X^n, \hat{f}}, z \models \psi$ and for counterfactual antecedents $\phi_1, \phi_2 , \ldots, \phi_n$ describing features of $X^n$ such that ${\cal C}_{X^n, \hat{f}}, z \models \phi_1 \cftual \neg \psi,  \phi_2 \cftual \psi, \ldots, \phi_n \cftual \neg \psi$ such that $\phi_{i+1} \models \phi_i$ and $\phi_i \not\models \phi_{i+1}$, there are collinear $x_1,... x_n \in X^n$ such that for each $i$, $x_i$  is a closest point to $z$  such that ${\cal C}_{X^n, \hat{f}}, x_i \models \phi_i$  and $||x_{i+1}-z|| > ||x_{i} -z||$.
\end{definition}
\begin{rem}
An L1 norm for a counterfactual model is a logical specificity respecting norm. 
\end{rem}

In addition, a flip (move from a point verifying $\phi_1$ to a point verifying $\phi_2$ corresponds to a move from a transformation $\Delta_i$ to a transformation $\Delta_j$ with $i \subset j$.  Thus, flips determine a partial ordering under $\subseteq$ over the shifted dimensions $i$: thus  $\Delta_i \leq \Delta_j$, if $i \subseteq j$. We are interested in the behavior of $\hat{f}$ with respect to the partial ordering on $\Delta_i$.

\begin{definition}$\hat{f}$ is nearly constant around $\x$, if for every sufficiently minimally appropriate $\Delta_i$
for all $\Delta_j \supset \Delta_i$, $\hat{f}(\Delta_j (\x)) = \hat{f}(\Delta_i (\x))$.
\end{definition}
A nearly constant $\hat{f}$  changes  values only once for each combination of features/dimensions $d_i$ moving out from a focal point $\x$.  So at some distance $d$, nearly constant $\fh$ becomes constant $\fh$.  For a nearly constant $\hat{f}$ around $\x$, ${\bf S}_{\x, \pi}$, has flip degree 1.  A complete local explanation for  $\fh$'s prediction of $\pi$  within $d$, ${\bf S}^d_{\x, \pi}$, is a global explanation $\hat{f}$'s behavior with respect to $\pi$.  

We can generalize this notion to define an  $n${\em -shifting} $\hat{f}$.  If $\hat{f}$ flips values $n$ times moving out from $\x$, ${\bf S}_{\x, \pi}$ has flip degree $n$.

 

\begin{proposition}  Suppose  A counterfactual model has a logical specificity respecting norm, then: ${\bf S}_{\x, \pi}$, has a flip degree $\leq 2$  iff  ${\cal N}_{\hat{f}, \x, \pi}$ forms a convex subspace of $\hat{f}[X]$.
\end{proposition}
\begin{sketch} 
Assume ${\bf S}_{\x, \pi}$ has flip degree  $\geq 3$.  Then ${\bf S}_{\x, \pi}$ will contain counterfactuals with antecedents $\phi$, $\chi$, $\delta$ such that $\phi \models \chi \models\delta$ but, say, $\phi$ and $\delta$ counterfactually support $\pi$ but not $\chi$.  As the underlying norm respects $\models$, there are collinear points $x$, $y$, and $z$, where $x$ is a closest point to $\x$ where $\phi$ is true,  y is a closest $\chi$ world, and z is a closest $\delta$ world such that $\x < z < y < x$.  But $\x, y_\chi \in {\cal N}_{\hat{f}, \x, \pi}$, while $x_\phi, z_\delta \in {\bf B}_{\x,\pi}$ and $\not\in {\cal N}_{\hat{f}, \x, \pi}$, which makes ${\cal N}_{\hat{f}, \x, \pi}$ non convex.  Conversely, suppose  ${\cal N}_{\hat{f}, \x, \pi}$ is non convex.  Using the construction of counterfactuals from the boundary ${\bf B}_{\x,\pi}$ of  ${\cal N}_{\hat{f}, \x, \pi}$ will yield a set  with flip degree 3 or higher. 
\end{sketch}

The flip degree of ${\bf S}_{\x, \pi}$ gives a measure of the degree of non-convexity of  ${\cal N}_{\hat{f}, \x, \pi}$, and a measure of the complexity of an explanation of $\fh$'s behavior.  A low flip degree for ${\bf S}^d_{\x, \pi}$ with minimal overdeterminations provides a more general and comprehensive explanation.  With Proposition 4, a low flip degree converts a local complete explanation into a global explanation, which is  {\em a priori} preferable.   It is also arguably closer to our prior beliefs about basic causal processes.   The size of ${\bf S}^d_{\x, \pi}$ gives us a measure to evaluate $\fh$ itself; a large ${\bf S}^d_{\x, \pi}$ doesn't approximate very well a good scientific theory or the causal structures postulated by science.  Such a $\fh$
 lacks generality; it has neither captured the sufficient nor the necessary conditions for its predictions in a clear way. This could be due to a bad choice of features determining $\fh$'s input $X^n$ \cite{dube:2018}; too low level or unintuitive features could lead to lack of generality with high flip degrees and numerous overdeterminations.  Thus, we can use ${\bf S}^d_{\x, \pi}$ to evaluate $\fh$ and its input representation $X^n$.  
 
 The flip degree of ${\bf S}_{\x, \pi}$ and the topology of ${\cal N}_{\hat{f}, \x, \pi}$ can also tell us about the relation between counterfactual explanations based on some element in $X$ and ground truth instances provided during training.  Our learning algorithm $\fh$ is trying to approximate or learn some phenomenon, which we can represent as a function $f: X \rightarrow Y$; the observed pairs $(z, f(z))$ are ground truth points for $\fh$.  Ideally, $\fh$ should fit and converge to $f$---i.e., with the number of data points $N$ $\fh$ is trained on $lim_{N \rightarrow \infty} \fh^N \rightarrow f $; in the limit explanations of the behavior of $\fh$ will explain $f$, the phenomenon we want to understand.   Given that we generate counterfactual situations using techniques used to find adversarial examples, however, counterfactual explanations may also be based on adversarial examples that have little to no intuitive connection with the ground truth instances $\fh$ was trained on.  While these can serve to explain the behavior of $\hat{f}$ and as such can be valuable, they  typically aren't good explanations of the phenomenon $f$ that $\hat{f}$ is trying to model.   \cite{laugel:etal:2019} seek to isolate good explanations of $f$ from the behavior of $\fh$ and propose a criterion of topological connectedness for good counterfactual explanations.  This idea readily be implemented as a constraint on ${\cal N}_{\hat{f}, \x, \pi}$: roughly, $\hat{f}$ as an approximation of $f$ will yield good counterfactual explanations relative to a focal point $\x$ only if for any point $y$ outside of ${\cal N}_{\hat{f}, \x, \pi}$, there is a region $C$ where $\fh$ returns the same value and a path of points $y_1,... y_n \in C$ between $y$ and a ground truth data point $p$ such that $f(p) = \fh(p) = \fh(y_i) = \fh(y)$.\footnote{We note that our discussion and constraint make clear the distinction between $f$ and $\fh$ which is implicit in \cite{laugel:etal:2019,holzinger:etal:2020}.}




\section{Pragmatic constraints on explanations}

While we have clarified the partiality of counterfactual explanations, AI applications can encode data via hundreds even thousands of features.   Even for our simple running example of a bank loan program, the number of parameters might provide a substantial set of counterfactuals in the complete local explanation given by ${\bf S}_{\x, \pi}$.  This complete local explanation might very well involve too many counterfactuals for humans to grasp. We  still to understand what counterfactual explanations are {\em pragmatically relevant} in a given case.  

Pragmatic relevance relies on two observations.  First, once we move out a certain distance from the focal point, then the counterfactual shifts intuitively cease to be about the focal point; they cease to be counterparts of $\x$ and become a different case.  Exactly what that distance is, however, will depend on a variety of factors about the explainee ${\cal E}$ and what the explainer believes about ${\cal E}$.   Second, appropriate explanations must respond to the particular {\em conundrum} or cognitive problem that led ${\cal E}$ to ask for the explanation \cite{bromberger:1962,achinstein:1980,miller:2019}.   On our view, the explainee ${\cal E}$ requires an explanation when her beliefs do not lead her to expect the  observed prediction $\pi$.  When ${\cal E}$'s beliefs suffice to predict $\fh(\x) = \pi$, she has {\em a priori} an answer to the question {\em Why did $\fh(x) = \pi$}?  In our bank example from Section 2,  had $\cal{E}$'s beliefs been such that she did not expect a loan from the bank, she wouldn't have needed to ask, {\em why did the bank not give me a loan?}\footnote{Of course ${\cal E}$ might want to know whether her beliefs matched the bank's reasons for denying her a loan, but that's a different question---and in particular it's not a {\em why} question.}  

The conundrum comes from a mismatch between $\cal{E}$'s understanding of what $\fh$ was supposed to model (our function $f$) and $\fh$'s actual predictions.   So ${\cal E}$, in requesting an explanation of $\fh$'s behavior, might also want an explanation of $f$ itself (see the previous section for a discussion).  Either $\cal{E}$ is mistaken about the nature of $\hat{f}$, or her grasp of $\fh$ is incomplete..\footnote{Perhaps ${\cal E}$ is also mistaken about or has an incomplete grasph of $f$ or if not, she is mistaken about how $\hat{f}$ differs from $f$).  But we will not pursue this here.}  More often than not, $\cal{E}$ will have certain preconceptions about $\fh$, and then many if not most of the counterfactuals in ${\bf S}_{\x, \pi}$ may be irrelevant to ${\cal E}$.   
  A relevant or fair and adequate explanation for $\cal{E}$ should provide a set $\mathfrak{C}^d_{\cal E}$ of appropriate $\Delta_i$ with $\|\Delta_i(\x) -\x\| \leq d$ showing which of $\cal{E}$'s assumptions were faulty or incomplete, thus solving her conundrum.   

Suppose that the explainee $\cal{E}$ requests an explanation why $\hat{f}(\x) = \eta$, and that  $\x$ is decomposed into $\langle x_{\vec{d_1}}, x_{\vec{d_2}}\rangle$.
\begin{enumerate}
\item [CI] Suppose $\cal{E}$'s conundrum based on incompleteness; i.e., the conundrum arises from the fact that for $\cal{E}$ $\fh$ only pays attention to the values of dimensions $\vec{d_1}$ in the sense that for her $\fh(\langle x_{\vec{d_1}}, x_{\vec{d_2}}\rangle) = \fh(\langle x_{\vec{d_1}}, x'_{\vec{d_2}}\rangle)$, for any values  $x'_{\vec{d_2}}$.  Then there is a $\Delta \in \mathfrak{C}^d_{\cal E}$ such that $\Delta(\langle x_{\vec{d_1}}, x_{\vec{d_2}}\rangle) = \langle x_{\vec{d_1}}, y_{\vec{d_2}}\rangle$ and $\hat{f}(\Delta(\x)) = \hat{f}(\langle x_{\vec{d_1}}, y_{\vec{d_2}}\rangle) = \pi$ while $\hat{f}(\x) = \fh(\langle x_{\vec{d_1}}, x_{\vec{d_2}}\rangle) = \eta.$
\item [CM] Suppose $\cal{E}$'s conundrum is based on a mistake.   
Then there is a $\Delta \in \mathfrak{C}^d_{\cal E}$ such that $\Delta(\langle x_{\vec{d_1}}, x_{\vec{d_2}}\rangle) = \langle y_{\vec{d_1}}, x_{\vec{d_2}}\rangle$ such that $\hat{f}(\langle y_{\vec{d_1}}, x_{\vec{d_2}}\rangle) = \hat{f}(\Delta(\x)) = \pi$.  I.e., $\Delta$ must resolve $\cal{E}$'s  conundrum by providing the values for the dimensions $\vec{d_2}$ of $\x$ on which $\cal{E}$ is mistaken.
\end{enumerate}


A fair and adequate explanation must not only contain counterfactuals that resolve the explainee's conundrum.  It must make clear the biases of the system which may account for $0$'s incomplete understanding of $\hat{f}$; it must lay bare any prejudicial factors $P$ that affect the explainee and thus in effect all overdetermining factors as in Definition \ref{def:over}. 
An explainee might reasonably want to know whether such biases resulted in a prediction concerning her.  E.g., the explanation in \ref{bankcounterfactual} might satisfy CM or CI, but still be misleading.  Thus:
\begin{enumerate}
\item [CB] 
$\forall$ prejudicial factors $P$, there is a $\Delta \in \mathfrak{C}^d_{\cal E}$ such that $\hat{f}(\Delta(\x)) = \pi$ and $P(\Delta(\x))= \Delta(\x)$.
\end{enumerate}
In our bank loan example, if the bank is constrained to provide an explanation obeying CB, then it must provide an explanation according to which being white and having ${\cal E}$'s salary would have sufficed to get the loan. 
\begin{definition} \label{fair} A set of counterfactuals provides a  {\em fair and adequate} explanation of $\fh$ for ${\cal E}$ at $\x$ just in case they together satisfy CM, CI and CB within a certain distance $d$ of $\x$.
\end{definition}   
The counterfactuals in $\mathfrak{C}^d_{\cal E}$ jointly provide a {\em fair and adequate} explanation of $\fh$ for ${\cal E}$, though individually they may not satisfy all of the constraints.   We investigate how hard it is to find an adequate local explanation in the next section.

\section{The algorithmic complexity of finding fair and adequate explanations}
In this section, we examine the computational complexity of finding a fair and adequate explanation.  To find an appropriate explanation, we imagine a game played, say, between the bank and the would-be loan taker $\cal{E}$ in our example from Section 2, in which $\cal{E}$ can ask questions of the bank (or owner/ developer of the algorithm) about the algorithm's decisions.  We propose to use a two player game, an {\em explanation game} to get appropriate explanations for the explainee.

The pragmatic nature of explanations already motivates the use of a game theoretic framework. We have argued fair and adequate explanations must obey pragmatic constraints; and in order to satisfy these in a cooperative game the explainer must understand explainee ${\cal E}$'s conundrum and respond so as to resolve it. Providing an explanation is a pragmatic act that  takes  into  account  an  explainee’s  cognitive  state  and the  conundrum it engenders for the particular fact that needs explaining.   A  cooperative  explainer  will provide  an  explanation  in  terms  of  the type he assigns to ${\cal E}$, as the type will encode the relevant portions of   ${\cal E}$’s cognitive state.  On the other hand the explainee will need to interpret the putative explanation in light of her model of the explainer’s view of his type.  Thus, both explainer and explainee  naturally  have  strategies  that  exploit  information about the other.  Signaling games \cite{spence:1973} are a well-understood and natural formal framework in which to explore the interactions between explainer and explainee; the game theoretic machinery we develop below can be easily adapted into a signaling game between explainer and explainee where explanations succeed when their strategies coordinate on the same outcome. 

Rather than develop signaling games however for coordinating on successful explanations, we look at non-cooperative scenarios where the explainer $\fh$ may attempt to hide a good explanation.  For instance, the bank in our running example might have encoded directly or indirectly biases into its loan program that are prejudicial to $\cal{E}$, and it might not want to expose these biases. The games below provide a formal account of the difficulty our explainee has in finding a winning strategy in such a setting.

To define an explanation game, we first fix a  set of {\sf  two players}  $\{{\cal E},{\cal A}\}$. 
  
The moves or actions $V_{\cal E}$ for explainee ${\cal E}$ are: playing an ACCEPT move---in which ${\cal E}$ accepts a proposed $\Delta_i$ if it partially solves her conundrum; playing an N-REQUEST move---i.e. requesting a $\Delta_j$ where $j$ differs from all i such that $\Delta_i$ has been proposed by ${\cal A}$ in prior play; playing a P-REQUEST move---i.e. for some particular $i$, requesting $\Delta_i$.  $\cal{E}$ may also play a CHALLENGE move, in which ${\cal E}$ claims that a set of features $A_1, \ldots A_n$ of the focal point that entails $\pi$ in the counterfactual model associated with $\fh$.
We distinguish three types of ME explanation games for ${\cal E}$ based on the types of moves she is allowed: the {\em Forcing} ME explanation games, in which ${\cal E}$ may play ACCEPT, N-REQUEST, P-REQUEST; the more restrictive {\em Restriction} ME explanation games, in which ${\cal E}$ may only play ACCEPT, N-REQUEST; and finally Challenge ME explanation games in which CHALLENGE moves are allowed.

Adversary ${\cal A}$'s moves $V_{\cal A}$ consists of the following: producing $\Delta_i$ and computing $\fh(\Delta_i(\x))$ in response to N-REQUEST or P-REQUEST by $\cal{E}$; if $\game$ is a forcing game, ${\cal A}$ must play $\Delta_i$ at move $m$ in $\play$, if $\cal{E}$ has played P-REQUEST $\Delta_i$ at $m-1$.  In reacting to a N-REQUEST, player ${\cal A}$ may offer any new $\Delta_i$; if he is noncooperative, he will offer a new $\Delta_i$ that is not relevant to $\cal{E}$'s conundrum, unless he has no other choice.  On the other hand, ${\cal A}$ must react to a CHALLENGE move by $\cal{E}$ by playing a $\Delta_i$ that either completes or corrects the Challenge assumption.  A CHALLENGE demands a cooperative response; and since it can involve any implicitly definable prejudicial factor as in Definition 5, it can also establish CB, as well as remedy CI or CM.    





We now specify a win-lose, generic {\em explanation game}. 
\begin{definition}
  An {\sf Explanation game}, $\game$, concerning a polynomially computable function $\hat{f}\colon  X^n \rightarrow Y$, where $X^n$ is a space of boolean valued features for the data and $Y$ a set of predictions, is a tuple $((V_{\cal E} \cup V_{\cal A})^*, {\cal E}, {\cal A}, \hat{f}\colon  X^n \rightarrow Y, \x, d, \mathfrak{C}^d_{\cal E})$ where: 
  \begin{enumerate}
    
            \item [i.] $\mathfrak{C}^d_{\cal E} \subseteq {\bf B}^d_{\x, \pi}$ resolves ${\cal E}$'s conundrum and obeys CB. 
      \item [ii.] $\x \in X^n$ is the starting position, $d$ is the antecedently fixed distance parameter.
        \item [iii.] ${\cal A}$, but not $\cal{E}$ has access to the behavior of $\hat{f}$ and {\em a fortiori} $\mathfrak{C}^d_{\cal E}$.
      \item [iv.] $\cal{E}$ opens $\game$ with a REQUEST or CHALLENGE move 
      \item [v.] ${\cal A}$ responds to $\cal{E}$'s requests by playing some $\Delta_i, i \leq d$.
      \item [vi.] $\cal{E}$ may either play ACCEPT, in which case the game ends or again play a REQUEST or CHALLENGE move.

      \end{enumerate}
\end{definition}
\noindent
$\cal{E}$ {\em wins} $\game$ just in case in $\game$ she can determine $\mathfrak{C}^d_{\cal E}$.  The game terminates when (a) 0 has determined $\mathfrak{C}^d_{\cal E}$ (resolved her conundra) or gives up. 

$\cal{E}$ always has a winning strategy in an explanation game.  The real question is how quickly $\cal{E}$ can compute her winning condition.   An answer depends on what moves we allow for $\cal{E}$ in the Explanation game; we can restrict $\cal{E}$ to playing a Restriction explanation game, a Forcing game or  a Forcing game with CHALLENGE moves.  
\begin{proposition} \label{prop:pls}
Suppose $\game$ is a forcing explanation game.  Then the computation of $\cal{E}$'s winning strategy in $\game$ is Polynomial Local Search complete (PLS) \cite{johnson:etal:1988,papadimitriou:etal:1990}.  On the other hand if $\game$ is only a Restriction game, then the worst case complexity for finding her strategy is exponential.
\end{proposition}
\vspace{-0.15cm}
\begin{sketch}Finding $\mathfrak{C}^d_{\cal E}$ is a search problem using $\fh$.  $\mathfrak{C}^d_{\cal E}$ is finite with, say, $m$ elements.  These elements need not be unique; they just need jointly to solve the conundrum.  This search problem is PLS just in case every solution element is polynomially bounded in the size of the input instance, $\fh$ is poly-time, the cost of the solution is poly-time and it is possible to find the neighbors of any solution in poly-time. Let $\x$ be the input instance.  By assumption, $\hat{f}$ is polynomial; and given the bound $d$, the solutions $y$ for $\fh(y) = \pi$ and $y \in \mathfrak{C}^d_{\cal E}$  are polynomially bounded in the size of the description of $\x$.  Now, finding a point $y \in \mathfrak{C}^d_{\cal E}$ that solves at least part of ${\cal E}$'s conundrum, 
as well as finding neighbors of $y$ is poly-time, since $\cal{E}$ can use P-REQUEST moves to direct the search.  To determine the cost $c$ of finding $\mathfrak{C}^d_{\cal E}$ for $|\mathfrak{C}^d_{\cal E}| = m$ in poly-time: we set for $y \in \mathfrak{C}^d_{\cal E}$ the $j$th element of $\mathfrak{C}$ computed as $c(y) =  m-j$; if $y \not\in \mathfrak{C}$, $c(y) = m$.  Finding $\mathfrak{C}^d_{\cal E}$ thus involves determining $m$ local minima and is PLS.  In addition, determining $\mathfrak{C}^d_{\cal E}$ encodes the PLS complete problem FLIP \cite{johnson:etal:1988}: the solutions $y$ in $\game$ have the same edit distance as the solutions in FLIP, $\fh$ encodes a starting position, and our cost function can be recoded over the values of the Boolean features defining $y$ to encode the cost function of FLIP and the function that compares solutions in FLIP is also needed and constructible in $\game$.  So finding $\mathfrak{C}^d_{\cal E}$ is PLS complete in $\game$ as it encodes FLIP.

The fact that forcing explanation games are PLS complete makes getting an appropriate explanation computationally difficult.  Worse, 
if $\game$ is a Restriction Explanation game,
then ${\cal A}$ can force ${\cal E}$ to enumerate all possible $\Delta_i$ within radius $d$ of $\x$ to find $\mathfrak{C}^d_{\cal E}$. 
\end{sketch}

\begin{proposition} \label{prop:challenge}
Suppose $\game$ is a Challenge explanation game.  Then $\cal{E}$ has a winning strategy in $\game$ that is linear time computable.  
\end{proposition}
\vspace{-0.15cm}
\begin{sketch} ${\cal A}$ must respond to $\cal{E}$'s CHALLENGE moves by correcting or completing $\cal{E}$'s proposed list of features.  $\cal{E}$ can determine $\mathfrak{C}^d_{\cal E}$ in a number of moves that is linear in the size of $\mathfrak{C}^d_{\cal E}$.
\end{sketch}

A Challenge explanation game mimics a coordination game where ${\cal A}$ has perfect information about $\mathfrak{C}^d_{\cal E}$, because it forces cooperativity and coordination on the part of ${\cal A}$.  Suppose ${\cal E}$ in our bank example claims that her salary should be sufficient for a loan.   In response to the challenge, the bank could claim the salary is not sufficient; but that's not true---the salary {\em is} sufficient {\em provided} other conditions hold.  That is, ${\cal E}$'s conundrum is an instance of CI.  Because of the constraint on CHALLENGE answers by the opponent, the bank must complete the missing element: {\em if you were white with a salary of \euro 50K,...}  Proposition \ref{prop:challenge} shows that when investigating an $\fh$ in a challenge game, exploiting a conundrum is a highly efficient strategy.     

The flip degree of ${\bf S}^d_{\x, \pi}$ and the number of overdetermining factors $O(x, \pi)$ (Definition \ref{def:over}) typically affect the size of $\mathfrak{C}$ and thus the complexity of the conundrum and search for fair and adequate explanations and their logical valid associates.  More particularly, when $| O(\pi,\x) | = n$ and the cost of the prediction is as in the proof of Proposition \ref{prop:pls},  $\cal{E}$'s conundrum and the explanations resolving it may require $n$ local minima.  When the flip degree of ${\bf S}^d_{\x, \pi}$ is $m$, $\cal{E}$ may need to compute $m$ local minima.

To develop practical algorithms for fair and adequate explanations for AI systems, we need to isolate ${\cal E}$'s conundrum.  This will enable us to exploit the efficiencies of Challenge explanation games.   Extending the framework to discover ${\cal E}$'s conundrum behind her request for an explanation is something we plan to do using epistemic games from \cite{JOLLI} with more developed linguistic moves.  In a more restricted setting where Challenge games are not available, our game framework shows that clever search algorithms and heuristics for PLS problems will be essential to providing users with relevant, and provably fair and adequate counterfactual explanations.  This is something current techniques like enumeration or finding closest counterparts, which may not be relevant \cite{marques-silva:etal:2019,ignatiev:etal:2020,karimi:etal:2020}---do not do.    

\vspace{-0.3cm}
\section{Conclusion}

We have shown that counterfactual explanations can deliver partial, but epistemically accessible and adequate explanations.  We have also shown that any counterfactual explanation can be extended to a valid deductive one.  We have shown that pragmatic factors dramatically affect the complexity of finding adequate explanations, and we introduced Explanation Games, which provided  to represent finding fair and adequate counterfactual explanations as a PLS complete search problem.  In addition, we explored how the complexity of the set of counterfactuals describing a local neighborhood around the focal point can affect both the complexity of fair and adequate explanations and our evaluation of the learning algorithm as a model. 

Our paper fills in part of the gap for finding fair and adequate explanations in a computationally reasonable way.  Nevertheless moving from an explanation provided by an explanation game to a proof from a minimal set of sufficient premises as in Proposition \ref{cftual-proof}  is still computationally difficult.  In future work we will look at efficient heuristics for this step.  In future work, we will alo look at how explanation games help us to formally explore interactive machine learning, in particular ``human in the loop'' or interactive explainability for machine learning function behavior \cite{amershi:etal:2014,holzinger:etal:2019}.   Such game theoretic investigations may have special relevance in medical domains \cite{holzinger:etal:2021}.  




\section*{Acknowledgement}We thank the ANR PRCI grant SLANT, the ICT 38 EU grant COALA and the 3IA Institute ANITI funded by the ANR-19-PI3A-0004 grant for research support. We alo thank the reviewers for their insightful comments.

\bibliographystyle{splncs04}
\bibliography{cntfls,erc2}

\begin{thebibliography}{10}
\providecommand{\url}[1]{\texttt{#1}}
\providecommand{\urlprefix}{URL }
\providecommand{\doi}[1]{https://doi.org/#1}

\bibitem{achinstein:1980}
Achinstein, P.: The Nature of Explanation. Oxford University Press (1980)

\bibitem{amershi:etal:2014}
Amershi, S., Cakmak, M., Knox, W.B., Kulesza, T.: Power to the people: The role
  of humans in interactive machine learning. Ai Magazine  \textbf{35}(4),
  105--120 (2014)

\bibitem{JOLLI}
Asher, N., Paul, S.: Strategic conversation under imperfect information:
  epistemic {M}essage {E}xchange games. {L}ogic, {L}anguage and {I}nformation
  \textbf{27.4},  343--385 (2018)

\bibitem{bachoc:etal:2018}
Bachoc, F., Gamboa, F., Halford, M., Loubes, J.M., Risser, L.: Entropic
  variable projection for explainability and intepretability. arXiv preprint
  arXiv:1810.07924  (2018)

\bibitem{bromberger:1962}
Bromberger, S.: An approach to explanation. In: Butler, R. (ed.) Analytical
  Philsophy, pp. 72--105. Oxford University Press (1962)

\bibitem{chang:keisler:1973}
Chang, C.C., Keisler, H.J.: Model theory. Elsevier (1990)

\bibitem{deraedt:etal:2020}
De~Raedt, L., Duman{\v{c}}i{\'c}, S., Manhaeve, R., Marra, G.: From statistical
  relational to neuro-symbolic artificial intelligence. arXiv preprint
  arXiv:2003.08316  (2020)

\bibitem{doshi:etal:2017}
Doshi-Velez, F., Kim, B.: Towards a rigorous science of interpretable machine
  learning. arXiv preprint arXiv:1702.08608  (2017)

\bibitem{dube:2018}
Dube, S.: High dimensional spaces, deep learning and adversarial examples.
  arXiv preprint arXiv:1801.00634  (2018)

\bibitem{fan:toni:2015a}
Fan, X., Toni, F.: On computing explanations in argumentation. In: Bonet, B.,
  Koenig, S. (eds.) Proceedings of the Twenty-Ninth {AAAI} Conference on
  Artificial Intelligence. pp. 1496--1502. {AAAI} Press (2015)

\bibitem{friedrich:zanker:2011}
Friedrich, G., Zanker, M.: A taxonomy for generating explanations in
  recommender systems. AI Magazine  \textbf{32}(3),  90--98 (2011)

\bibitem{gardenfors:makinson:1988}
G\"ardenfors, P., Makinson, D.: Revisions of knowledge systems using epistemic
  entrenchment. In: Vardi, M.Y. (ed.) Proceedings of the Second Conference on
  Theoretical Aspects of Reasoning about Knowledge. pp. 83--95. Morgan
  Kaufmann, San Francisco (1988)

\bibitem{ginsberg:1986}
Ginsberg, M.L.: Counterfactuals. Artificial intelligence  \textbf{30}(1),
  35--79 (1986)

\bibitem{hempel:1965}
Hempel, C.G.: Aspects of scientific explanation. Free Press New York (1965)

\bibitem{holzinger:etal:2020}
Holzinger, A., Carrington, A., M{\"u}ller, H.: Measuring the quality of
  explanations: the system causability scale (scs). KI-K{\"u}nstliche
  Intelligenz pp.~1--6 (2020)

\bibitem{holzinger:etal:2021}
Holzinger, A., Malle, B., Saranti, A., Pfeifer, B.: Towards multi-modal
  causability with graph neural networks enabling information fusion for
  explainable ai. Information Fusion  \textbf{71},  28--37 (2021)

\bibitem{holzinger:etal:2019}
Holzinger, A., Plass, M., Kickmeier-Rust, M., Holzinger, K., Cri{\c{s}}an,
  G.C., Pintea, C.M., Palade, V.: Interactive machine learning: experimental
  evidence for the human in the algorithmic loop. Applied Intelligence
  \textbf{49}(7),  2401--2414 (2019)

\bibitem{ignatiev:etal:2020}
Ignatiev, A., Narodytska, N., Asher, N., Marques-Silva, J.: On relating
  “why?” and “why not?” explanations. In: Proceedings of AI*IA 2020
  (2020)

\bibitem{marques-silva:etal:2019}
Ignatiev, A., Narodytska, N., Marques-Silva, J.: On relating explanations and
  adversarial examples. In: Advances in Neural Information Processing Systems
  (2019)

\bibitem{johnson:etal:1988}
Johnson, D.S., Papadimitriou, C.H., Yannakakis, M.: How easy is local search?
  Journal of computer and system sciences  \textbf{37}(1),  79--100 (1988)

\bibitem{junker:2004}
Junker, U.: Preferred explanations and relaxations for over-constrained
  problems. In: AAAI-2004 (2004)

\bibitem{karimi:etal:2020}
Karimi, A.H., Barthe, G., Balle, B., Valera, I.: Model-agnostic counterfactual
  explanations for consequential decisions. In: International Conference on
  Artificial Intelligence and Statistics. pp. 895--905. PMLR (2020)

\bibitem{kurakin:etal:2016}
Kurakin, A., Goodfellow, I., Bengio, S.: Adversarial examples in the physical
  world. arXiv preprint arXiv:1607.02533  (2016)

\bibitem{kusner:etal:2017}
Kusner, M.J., Loftus, J., Russell, C., Silva, R.: Counterfactual fairness. In:
  Advances in Neural Information Processing Systems. pp. 4066--4076 (2017)

\bibitem{laugel:etal:2019}
Laugel, T., Lesot, M.J., Marsala, C., Renard, X., Detyniecki, M.: Unjustified
  classification regions and counterfactual explanations in machine learning.
  In: Joint European Conference on Machine Learning and Knowledge Discovery in
  Databases. pp. 37--54. Springer (2019)

\bibitem{lewis:1973}
Lewis, D.: Causation. Journal of Philosophy  \textbf{70}(17),  556--567 (1973)

\bibitem{lundberg:lee:2017}
Lundberg, S.M., Lee, S.: A unified approach to interpreting model predictions.
  In: NIPS. pp. 4765--4774 (2017)

\bibitem{miller:2019}
Miller, T.: Explanation in artificial intelligence: Insights from the social
  sciences. Artificial Intelligence pp. 1--38 (2019)

\bibitem{molnar:2019}
Molnar, C.: Interpretable machine learning. Lulu. com (2019)

\bibitem{murdoch:etal:2019}
Murdoch, W.J., Singh, C., Kumbier, K., Abbasi-Asl, R., Yu, B.: Definitions,
  methods, and applications in interpretable machine learning. Proceedings of
  the National Academy of Sciences  \textbf{116}(44),  22071--22080 (2019)

\bibitem{papadimitriou:etal:1990}
Papadimitriou, C.H., Sch{\"a}ffer, A.A., Yannakakis, M.: On the complexity of
  local search. In: Proceedings of the twenty-second annual ACM symposium on
  Theory of computing. pp. 438--445 (1990)

\bibitem{pearl:1990}
Pearl, J.: System \textsc{Z}: a natural ordering of defaults with tractable
  applications to nonmonotonic reasoning. In: Proceedings of the 3rd conference
  on Theoretical aspects of reasoning about knowledge (TARK'90). pp. 121--135
  (1990)

\bibitem{peyre:2019}
Peyr{\'e}, G., et~al.: Computational optimal transport: With applications to
  data science. Foundations and Trends in Machine Learning  \textbf{11}(5-6),
  355--607 (2019)

\bibitem{ribeiro:etal:2016}
Ribeiro, M.T., Singh, S., Guestrin, C.: "why should {I} trust you?": Explaining
  the predictions of any classifier. In: KDD. pp. 1135--1144 (2016)

\bibitem{ribeiro:etal:2018}
Ribeiro, M.T., Singh, S., Guestrin, C.: Anchors: High-precision model-agnostic
  explanations. In: AAAI. pp. 1527--1535 (2018)

\bibitem{salzberg:1991}
Salzberg, S.: Distance metrics for instance-based learning. In: International
  Symposium on Methodologies for Intelligent Systems. pp. 399--408. Springer
  (1991)

\bibitem{spence:1973}
Spence, A.M.: Job market signaling. Journal of Economics  \textbf{87}(3),
  355--374 (1973)

\bibitem{wachter:etal:2017}
Wachter, S., Mittelstadt, B., Russell, C.: Counterfactual explanations without
  opening the black box: Automated decisions and the gpdr. Harv. JL \& Tech.
  \textbf{31}, ~841 (2017)

\bibitem{williamson:1988}
Williamson, T.: First-order logics for comparative similarity. Notre Dame
  Journal of Formal Logic  \textbf{29}(4) (1988)

\bibitem{younes:2018}
Younes, L.: Diffeomorphic learning (2019), arXiv.1806.01240

\end{thebibliography}


\end{document}